\documentclass[10pt, a4paper, conference, compsocconf]{IEEEtran}

\usepackage{graphicx}	
\usepackage{amsmath}
\graphicspath{/media/akshay/Stuff1/Selfie/Paper/graphics/}
\usepackage{multirow}
\usepackage{graphicx}
\usepackage{array}
\usepackage{enumerate}
\usepackage{amssymb}
\usepackage[tight,footnotesize]{subfigure}
\begin{document}
%
\title{Selfie Detection by Synergy-Constriant Based Convolutional Neural Network}

\author{\IEEEauthorblockN{Yashas Annadani, Vijaykrishna Naganoor, Akshay Kumar Jagadish and Krishnan Chemmangat     }
\IEEEauthorblockA{ Electrical and Electronics Engineering,
NITK-Surathkal, India.\\ }

}
\maketitle

\begin{abstract}

   Categorisation of huge amount of data on the multimedia platform is a crucial task.
In this work, we propose a novel approach to address the subtle problem of selfie detection for image database segregation on the web, given rapid rise in the number of selfies being clicked. A Convolutional Neural Network (CNN) is modeled to learn a synergy feature in the common subspace of head and shoulder orientation, derived from Local Binary Pattern (LBP) and Histogram of Oriented Gradients (HOG) features respectively. This synergy was captured by projecting the aforementioned features using Canonical Correlation Analysis (CCA). We show that the resulting network’s convolutional activations in the neighbourhood of spatial keypoints captured by SIFT are discriminative for selfie-detection. In general, proposed approach aids in capturing intricacies present in the image data and has the potential for usage in other subtle image analysis scenarios apart from just selfie detection. We investigate and analyse the performance of the popular CNN architectures (GoogleNet, Alexnet), used for other image classification tasks, when subjected to the task of detecting the selfies on the multimedia platform. The results of the proposed approach are compared with these popular architectures on a dataset of ninety thousand images comprising of roughly equal number of selfies and non-selfies. Experimental results on this dataset shows the effectiveness of the proposed approach. 
\end{abstract}


\begin{IEEEkeywords}
	Selfie; Deep Learning; Convolutional Neural Networks; Canonical Correlation Analysis
	
\end{IEEEkeywords}

%
\IEEEpeerreviewmaketitle

\section{Introduction}
 Self-Portraits, popularly known as Selfies have become ubiquitous over the past five years, with the burgeoning of social network and photo sharing platforms. Manually annotating the images as self portraits would be impossible given the massive volume of images entering the web. On the other hand, the tags that come along  with these images are either not available always or not reliable, if present. In this milieu, it becomes extremely important to design an efficient method that classifies a given image as selfie or not as  it has profound applications in sentiment analysis \cite{weiser2015me}~\cite{qiu2015does}, large scale image database segregation and retrieval~\cite{singhai2010survey}, and other psychological studies \cite{sorokowski2015selfie}~\cite{warfield2014making}. Apart from the aforementioned applications, it can also be used in scene understanding as people take selfies not just as a picture of oneself, but to capture the essence of the background, like a historic monument, that might exist in the background.
 
 We have hence formulated selfie detection as a binary classification problem where an image has to classified as a selfie or not a selfie and address this problem using Convolutional Neural Networks (CNN), which capture view invariant feature representations and have become successful in the recent past on many image classification tasks like object recognition~\cite{krizhevsky2012imagenet}~\cite{szegedy2015going}, object detection \cite{girshick2014rich} and face recognition \cite{parkhi2015deep}.
 \begin{figure}[!t]
 	\centering
 	
 	{  \includegraphics[width=\linewidth]{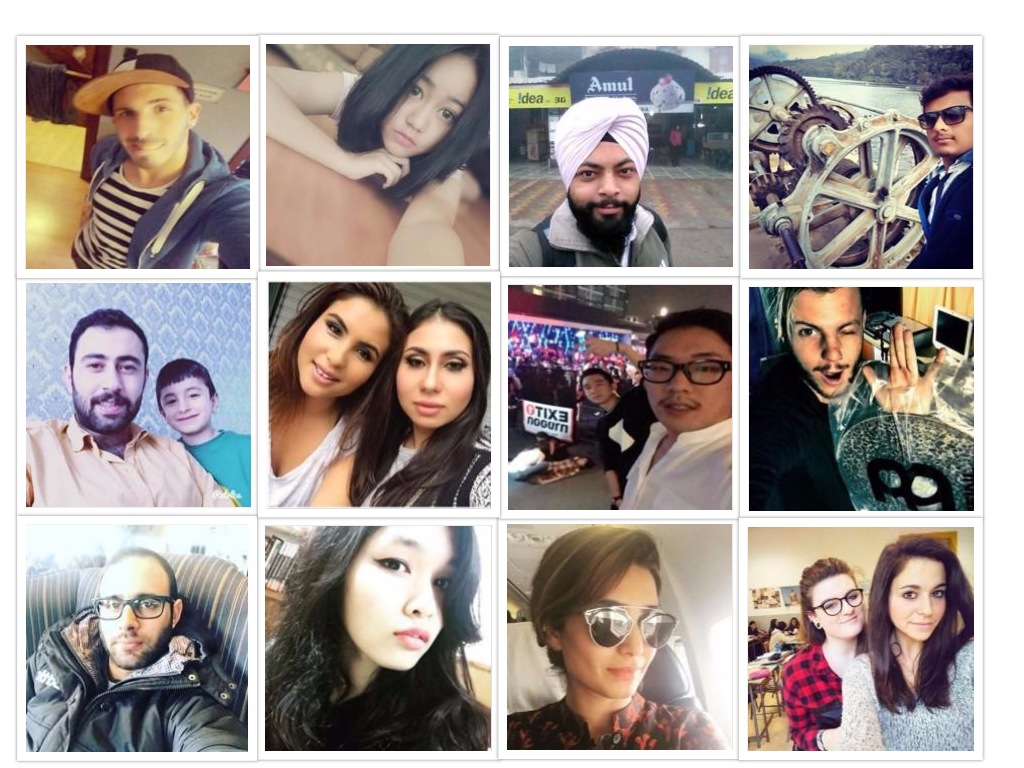}
 	}
 	
 	\caption{Sample images from the selfie dataset \cite{kalayeh2015take}.} 
 	\label{samplefig}
 \end{figure}
 
\begin{figure}[!t]
	\centering
	
	{  \includegraphics[width=\linewidth]{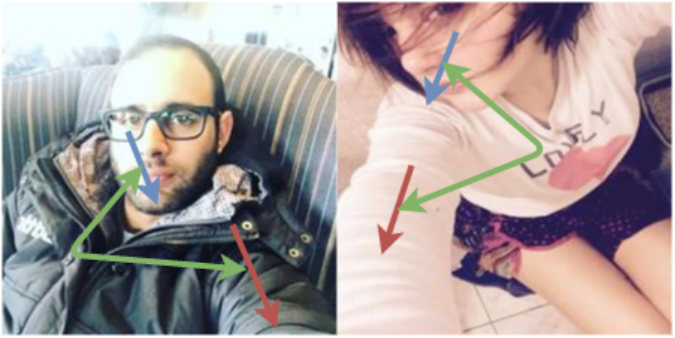}
	}
	
	\caption{An illustration of the synergy between the head orientation and shoulder-arm orientation in most of the selfies.} 
	\label{selfie_expl}
\end{figure}

Selfie detection is not a straightforward problem. To address this problem, we need to answer the following impending questions: What characterizes an image as a selfie? How do computer vision algorithms in the recent literature designed for other tasks perform on selfie detection as a surrogate task? Can we design and tract the algorithm to be able to scale it for larger and more subtle problems? An attempt has been made in this work to address these questions. In most of the cases, it is observed that humans tend to classify an image as selfie, by noticing the subtle poses of the self-portrait taker. For example, consider the selfie in Fig \ref{selfie_expl}, it can be easily seen that the strong visual cues for inferring a selfie is by making a connection between the shoulder-arm direction (indicated by the red vector) and the head gaze direction (indicated by the blue vector). A short-survey comprising of roughly 50 individuals, from diverse backgrounds, was conducted to verify if other individuals also share a similar opinion. It turned out that most of the people, 36 to be exact, considered hand and head orientation as a factor to classify or group the given images into  selfies or non-selfies. Thus, an attempt has been made in this regard and our entire approach consists of three stages- obtaining approximate head and shoulder orientation, followed by deriving the synergy or measure of similarity in their orientation and finally a constrained learning to model this synergy measure.

 To the best of our knowledge, there is no work in literature that exclusively addresses the problem of selfie-detection but there are a few that cater to the challenges faced in the three stages of our approach. There exists various types of approaches in literature that capture the head and shoulder orientation. Handcrafted approaches which involve using features like HOG, LBP, Histogram of Optical Flow (HOF) and other edge detection features are used to detect head and shoulders in images of people. Zheng \textit{et al.} \cite{zeng2010robust} used a multilevel Local Binary Pattern (LBP) \cite{ahonen2006face} to capture the head features and Histogram of Oriented Gradients (HOG) \cite{dalal2005histograms} for capturing the intensity gradient. Principal Component Analysis (PCA) was performed to capture only the essential components from these features and was applied for people counting in a scene. \cite{wang2013new} augmented the HOG-LBP feature with an edge feature which was learnt from different head shoulder edge contour proposals. Li \textit{et al.} \cite{li2008estimating} used a foreground segmentation algorithm along with the HOG feature to find head-shoulder. 
 
 These approaches have the advantage that they are computationally inexpensive while being rich enough to describe head and shoulders. But self-portraits of people occur in various poses, illumination and viewpoints, which require descriptors which are view-invariant and robust to changes in pose. Handcraft approaches do not offer these advantages, if used as standalone descriptors. Hence, we use HOG-LBP features to obtain only the alignment of head-shoulder, and not as a selfie-descriptor altogether. Higher level semantics are obtained by feature learning from Convolutional Neural Networks. Having obtained the head and shoulder orientation features, it is desirable to obtain the synergy measure between them. Canonical correlation analysis (CCA) is performed to project these features to a common subspace to find the synergy by virtue of their distance in this common subspace. CCA has been employed successfully in various cross modal matching algorithms for matching between images and text data \cite{hardoon2004canonical}, images and acoustic data \cite{izadinia2013multimodal} and in statistical applications where correlation between two random variables are to be maximized \cite{andrew2013deep}~\cite{malacarne2014canonical}. 

  As mentioned earlier, handcraft features fail as standalone descriptors when higher level of semantics to be learnt. Hence, a constrained CNN training scheme is employed such that merits of both are handcraft features and deep learning approach are utilised. This method of constraint learning was employed in face verification scenario in \cite{sarfraz2015deep}. A custom loss function was modeled in \cite{pathak2015constrained} for foreground segmentation, wherein the image-level tags constrained the CNN outputs. In our approach, the loss of CNN is modeled as least squared loss function between the network features and the synergy feature in the common subspace of HOG and LBP for head and shoulder respectively. Visualizations and results in Section \ref{expts} show that this method of training by making use of the handcrafted features is effective, thus rendering the model tractable. Besides, this also gives a better generalization of the approach beyond the selfie-detection problem. Finally, the constrained CNN is used as feature pools and features are extracted from the learnt convolutional maps at keypoints detected by SIFT \cite{lowe2004distinctive}. This approach is similar to \cite{wang2015action} where features are extracted at the dense trajectory locations. However, their approach does not employ any constraint based training paradigm. 
  
   In order to test the proposed algorithm, a dataset was compiled, 
 whose detailed description is presented in Section \ref{dataset}. The major contributions of this paper can be summarised as follows:
\begin{enumerate}
	\item Propose a novel approach which constraints a CNN to model handcrafted features in the common subspace obtained using CCA, making the method more tractable and scalable to situations involving higher complexities and 
	\item Demonstrate that the activations around the neighbourhood of SIFT keypoints when the network is trained in the above paradigm are highly discriminative for subtle tasks like selfie detection.
\end{enumerate}

The rest of the paper has been organized as follows, in Section \ref{dataset}, the dataset used for experimentation is described. In Section \ref{proposed}, proposed synergy-constraint learning based scheme for selfie detection is described, followed by experimental results in Section \ref{expts} and concluded by showing light on the future possibilities  in Section \ref{conclusion}.

\section{Dataset Description}
\label{dataset}
In \cite{kalayeh2015take}, a total of 85000 selfie images were collected from \textit{selfeed.com} using a real-time update of \#selfie on instagram and further manually annotated with different attributes such as age, gender and hair color. Doing so lead to elimination of 15,290 images. Further, images which were either completely irrelevant or general photographs of people were also removed. The resultant selfie dataset obtained predominantly has selfies containing single faces. In order to further diversify the dataset, additional images apart from the above were compiled from online and offline resources which mainly consisted of selfies having multiple people and also those clicked using equipments like selfie sticks. Thus the resultant dataset has one chunk of robust data (i.e. positives or selfies),  up-to-date images (i.e. positives or selfies)  with the current "selfie" trend.


Though negatives or non-selfies required for classification could have been compiled by arbitrarily collecting images not complying to the definition of selfie, by including images such as landscapes, animals and vehicles available in abundance, the augmented images were chosen with more specificity to present a more challenging and meaningful setting. It mainly comprises of images with at least one person performing a task (action dataset), posing for a picture (clicked by third person) and such. Some of the non-selfie images were obtained from Imagenet \cite{deng2009imagenet} database, under the hierarchies of people and objects, and INRIA Persons Dataset \cite{dalal2005histograms}, the rest were collected manually over time. Thus, a novel dataset that can serve as  good benchmark for testing algorithm was compiled.

\section{Proposed Approach}
\label{proposed}
As mentioned earlier, our approach shares the merits of both handcraft features and Deep Convolutional Neural Networks. The handcraft features extracted for constraint based CNN training paradigm and subsequent derivation of synergy feature for enforcing constraint on CNN has been elaborated below:

\subsection{Handcraft feature extraction:} Capturing the head features and shoulder alignment features forms the initial step of our algorithm. The approach of \cite{zeng2010robust} is employed, wherein multi-level Local Binary Pattern (LBP) gives the head alignment features and multi-level Histogram of Gradients (HOG) gives the head and shoulder alignment features. LBP is used as it is very successful for face and head detection tasks. But our framework is general and is applicable on other head features as well. In addition, HOG is computed as it is complementary to LBP \cite{ahonen2006face}, providing both head as well as shoulder alignment. In this section, we briefly revisit the feature extraction process in \cite{zeng2010robust}, which we have employed in our experiments.
All the images are converted to grayscale and resized to $227\times227$.

\subsubsection{Hierarchical HOG extraction} Histogram of Gradients (HOG)~\cite{dalal2005histograms} is a local descriptor capturing the intensity gradients. A  mask comprising of vector $[-1,0,1]$ is used to obtain the gradient magnitude. The resulting gradient magnitudes of the image is divided into blocks of $h=2^l \{l=1,2,3,\dots\}$, where $l$ is the level of the hierarchy. In our experiments, $l$ is chosen to be $4$. The gradient magnitudes are voted into $9$ bins of $0-180$ degrees of equal size separately from each of these blocks of images in every step of the hierarchy. Finally, the resulting histograms are concatenated to get the final descriptor $\gamma$.

\subsubsection{Hierarchical LBP extraction} Since the selfies mostly involve images of people taking picture of themselves, it is sensible to expect a face in the image. Local Binary Pattern (LBP) has been a very successful face descriptor \cite{shan2009facial} \cite{ahonen2006face}. Similar to HOG feature, the image is divided into blocks of $h=2^l\{l=1,2,3,\dots\}$. LBP patterns are obtained on each pixel corresponding to the neighbouring pixels of distance $d$. In our experiments, $h=8$, $d=1$. The LBP patterns are segregated into uniform and non-uniform patterns based on the occurrence of $0-1$ binary pairs. A pattern is considered as uniform pattern if the binary pattern contains a maximum of two $0-1$ transitions. Pixels in each block are voted into bins, such that each bin corresponds to a particular uniformity pattern. Non-uniform patterns are voted into a single bin. The resulting histograms are concatenated to get the final descriptor $\tau$.\\
\subsection{Synergy Feature Generation}
Based on the outcome of a short survey conducted, it was reaffirmed that majority of subjects usually perceive or classify an image as a selfie based on the head and shoulder position of the person clicking the selfie. It can also be observed that the head orientation and the shoulder-arm orientation of the person taking the selfie is usually in line with camera with his/her shoulder and head raised/pointing in the direction of the camera lens. While HOG and LBP features provide a good head and shoulder-arm orientation information, they do not provide any information about the synergy between these two. We hypothesize that capturing the synergy will ensure more discriminative features. In order to capture this, we propose to use canonical correlational analysis. 
 
 Canonical Correlation Analysis, popularly known as CCA, is a multivariate statistical technique that finds the linear relationship between two multidimensional datasets or features.  
 
 \begin{figure*}[!t]
	\centering
	
	{  \includegraphics[height=15 cm, width=\linewidth]{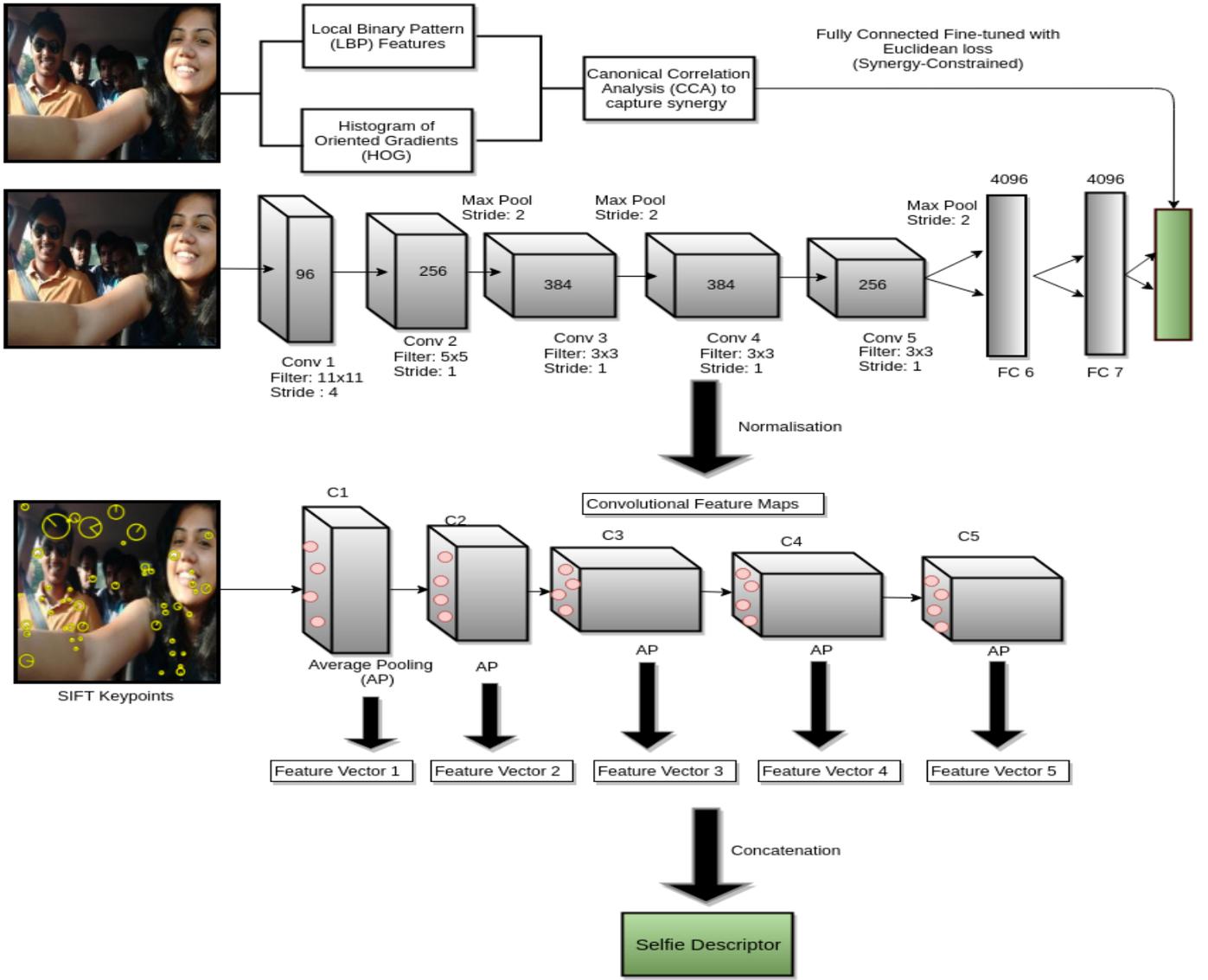}
	}
	
	\caption{A brief overview of the proposed approach} 
	\label{overall}\end{figure*}

 Let $\gamma~\in\mathbb{R}^c$ and $\tau~\in\mathbb{R}^d$ be the two multidimensional features (here, LBP and HOG) respectively. Then, CCA seeks to find the vectors ($a_{i},b_{i}$), known as canonical weights, that maximizes
 \begin{equation}
 \rho_{i}= corr(U_{i},V_{i})~\forall~i\in min\{c,d\} 
 \end{equation}
 where $U_{i} = a_{i}\gamma$ and $V_{i}=b_{i}\tau$. Thus, by maximizing the correlation between linear combination of variables of LBP feature extracted from every image and HOG feature variables extracted from the same image, $min\{c, d\}$ CCA modes (or canonical variates) uncorrelated with one another are obtained.
 
 In this scenario, CCA has been used as a tool that finds the best projection of the feature matrices onto a common subspace satisfying the aforementioned conditions, thereby resulting in two projection matrices, $U$ and $V$, of identical dimensions for each of the two extracted features. Hence synergy can be easily computed. Since these projection vectors have maximum correlation, we define synergy as the distance between these features in the common subspace. Mathematically, it is a vector which denotes the distance between the vectors. Therefore, synergy $S$ is:
 \begin{equation}
 \label{synergy}
 S=\dfrac{U-V}{||U-V||_2}
 \end{equation} 
 The resulting synergy feature $S$ is standardized to have unit variance and zero mean. Thus, the resulting synergy feature vector will have lower distance ($L2-Norm$) if the head and shoulder orientations are in synergy, otherwise it will have a higher distance.

\subsection{Constraint based CNN}
Given these handcrafted features, we aim to learn a multi-layer convolutional neural network architecture, which will take into consideration the synergy in the common subspace of these features. A Deep Convolutional Neural Network consists of a series of convolutional layers, interleaved by pooling and normalization layers. The convolutional layers are followed by fully-connected layers interleaved by dropout layers to prevent overfitting. These CNNs when trained end-to-end help in automated feature learning and classification. In this approach, we employ Alexnet architecture \cite{krizhevsky2012imagenet}, and use transfer learning by finetuning the model which has been pretrained on a large dataset of Imagenet \cite{deng2009imagenet}. This involves replacing the last fully connected layer by another fully connected layer of dimension equal to the synergy feature $S$. In addition, we pad the convolutional and pooling layers with size $(f-1)/2$, where $f$ is the filter size for the corresponding layer. This will help in preserving spatial dimensions thereby making the feature extraction from convolutional maps easier. In order to enforce the constraint on network to learn the synergy, the following loss function is modeled for the last fully connected layer: 
\begin{equation}
\ell=||\sigma(\Theta)-S||_2^2
\end{equation}
$\Theta$ is the vector of activations of the last modified fully-connected layer, $\sigma$ is \textit{ReLU} non-linearity followed by a soft-max operation and $S$ is the normalised synergy feature obtained through CCA. This euclidean norm will constraint the network to learn features in the convolutional layers which will be discriminative enough. The loss from the above equation is backpropogated and the entire network is retrained using the initial weights of Alexnet on Imagenet classification. Results shown in Section \ref{expts} further demonstrate that the convolutional activations learnt in this constrained manner are helpful for capturing intricacies.
\subsubsection*{Training Details}
To finetune the above network, stochastic gradient descent (SGD) is employed. The initial learning rate $\alpha$ is set to $10e^{-06}$ with a batch size of $16$. The learning rate is decreased by a factor of $0.5$ for every $2000$ iterations. We finetune the network for $8000$ iterations after which we did not observe any improvement in the loss on the validation set. 
\subsection{Extraction of selfie features from the constrained CNN} 
 After the training stage, this CNN is used as mere feature pools, where features are extracted at certain salient points from finetuned convolutional feature maps. We do not extract any features from the pooling layer as it is a subset of the previous convolutional layer when the pooling employed is max pooling. These salient points are obtained through SIFT \cite{lowe2004distinctive} keypoints. SIFT has been a popular keypoint detection and feature extraction algorithm for obtaining scale invariant fudicial points in an image. The motivation behind using SIFT keypoints is simple. Since selfies consist of images of varying degrees of rotation, scale and illumination, it is sensible to obtain SIFT keypoints which are robust to the aforementioned factors. For any given image $I$, let $\theta=(\{x_1,y_1\},\{x_2,y_2\},\dots,\{x_K,y_K\})$ be the location of keypoints obtained through SIFT. To extract features from an image, a forward pass is performed.  The corresponding convolutional activation maps are aggregated as $C=\{C_1,C_2,\dots,C_p,\dots,C_P\} $ with $C_p$ of dimensions $\mathbb{R}^{N_p\times W_p\times H_p}$ where $N_p$ is the number of filters in the $p^{th}$ convolutional layer, and $W_p$ and $H_p$ are the spatial resolutions of $p^{th}$ convolutional layer. Before extracting features from these learnt convolutional feature maps, they are normalised  as in \cite{wang2015action}. As the activations can vary greatly in number across filters in the same convolutional layer, the following normalisation is performed:
 \begin{equation}
 \hat{C_p}(n,x,y)=\dfrac{C_p(n,x,y)}{\max\limits_{n,x,y}|C_p(n,x,y)|}
 \end{equation}
 \begin{equation*}
  n=\{1,\dots,N_p\},x=\{1,\dots,W_p\},y=\{1,\dots,H_p\}
 \end{equation*}
 This normalisation ensures that all the activations are in the range $[-1,1]$ thereby reducing the undesirable influence of large activations. From this normalised feature pool, feature descriptor is obtained by averaging the max pool of activations in the 4-connected neighbourhood of all the keypoints. For tracking the keypoint locations across different convolutional maps of different sizes, we take into account the map size ratios, obtained by using the stride information from each of the layers. More specifically, for each convolutional feature map, the features are extracted as follows:
\begin{equation}
t_p^i=\dfrac{1}{K} \sum_{k=1}^{K}\hat{C_p}(i,\phi(\overline{r_p\times x_k}),\phi(\overline{r_p\times y_k}))
\end{equation}
where $\phi$ is the operator that returns the max pooling of the 4-connected neighbourhood of its input pixel locations, $\overline{(.)}$ is the rounding operation, $K$ is the number of key points corresponding to the image from which convolutional feature maps are to be extracted and $r_p$ is the map size ratio of the corresponding convolutional layer and is obtained as:
\begin{equation}
r_p=\prod_{j=1}^{p}\dfrac{1}{st_j}
\end{equation}
where $st_j$ is the stride value of the convolutional or pooling layer at level $j$.  For example, the first convolutional layer will have map size ratio of $\dfrac{1}{4}$ as its stride value is $4$,  the second convolutional layer will have map size ratio of $\dfrac{1}{4}$, as its stride is $1$ and so on. \\

Finally, the resulting feature vectors from all the layers are concatenated to get a single feature vector $T=\{t_1,t_2,\dots,t_P\}$, which is the selfie descriptor. 
For classification, off-the-shelf SVM with a linear kernel is employed.
 
\section{Experiments}
\label{expts}
\subsection{Experimental Settings}
All the experiments were performed on the selfie dataset as presented in section \ref{dataset}. A split of  $60\%$ of the data was used for training, $10\%$ for validation and the remaining data was used for testing. Each split has almost equal number of positive samples and negative samples. The CNN fine-tuning was performed using caffe~\cite{jia2014caffe} library. For classification using SVM for our approach, we let the regularization parameter $C$ of SVM to be one. We use VLFeat library \cite{vedaldi2010vlfeat} to extract the SIFT keypoints.
\subsection{Experimental Results}    

\subsubsection{Quantitative Analysis}
The result of our algorithm and comparisons of performance of different CNNs in the literature which have been very successful can be found in Table \ref{accuracy}. To further validate the efficacy of the method, we have compared our method against the simple case of using an SVM classifier on the synergy feature $S$ extracted using Equation \ref{synergy}. Therefore, it does not involve any use of CNN. For training the CNNs, we use the same strategy of finetuning, but replacing the last fully connected layer with two outputs for two classes. It can be found from Table \ref{accuracy} that our constrained model performs better as compared to other models. More so, the synergy feature and SVM approach gives only an accuracy of $52.4\%$, implying that features from CNN are more discriminative than just raw handcraft features. The Alexnet without the constraint underperforms as compared to the constrained model. It is also evident that GoogleNet \cite{szegedy2015going} performs only slightly better by $0.5\%$ than Alexnet in the unconstrained situation. Hence, the latter was employed in this paper for enforcing the synergy constraints given its simpler architecture and computational feasibility. The accuracy of the proposed method is better by almost $4\%$ and is a simple proof that capturing the synergy, and enforcing the constraint while learning, helps to better the selfie detection accuracy. 

\begin{figure}[!t]
	\centering
	
	{  \includegraphics[width=\linewidth]{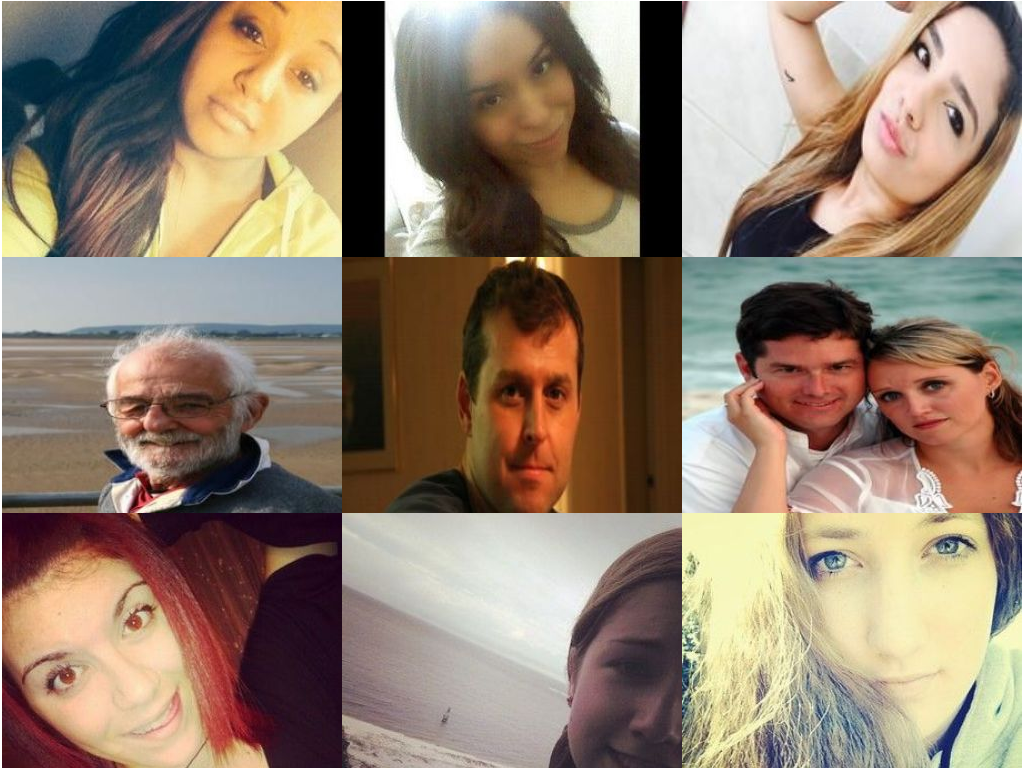}
	}
	
	\caption{Some of the true positives, false positives and false negatives of the proposed approach. Row 1 corresonds to true positves, Row 2 corresponds to false positives and Row 3 corresponds to false negatives.} 
	\label{tpfp}
\end{figure}

The top performing true positives, false negatives and false positives can be seen in Fig \ref{tpfp}. It can be observed that selfies are recognized based on  synergy even under different illumination levels, uncommon orientation and camera angles. The second row of the figure illustrates some of the false positives predicted by the approach. It can be seen that these images increasingly look like selfie, and is not easy even for a human to confidentially infer if those images are selfies. The third row indicates the false negatives. This is mainly due to the presence of just the face, which again renders the image ambiguous to the approach.\\
\begin{table}[]
	\centering
	\caption{Testing Accuracy in mAP on the augmented selfie dataset.}
	\label{accuracy}
	\begin{tabular}{|l|l|}
		\hline
	    \textbf{Model}& \textbf{Accuracy(mAP)}\\\hline
	    Synergy Feature + SVM & 52.4\%\\ \hline
		\textbf{Convolutional Networks} &\\ \hline
		
		Unconstrained AlexNet   \cite{krizhevsky2012imagenet}                                &          81.9\%       \\ \hline                                         
		Unconstrained  GoogleNet \cite{szegedy2015going}                                   & 82.4\%          \\ \hline
		\textbf{Synergy Constrained AlexNet (Proposed)} & \textbf{86.3\%} \\ \hline
	\end{tabular}
\end{table}

\begin{figure}[!b]
\centering
	{  \includegraphics[width=\linewidth, height=7 cm]{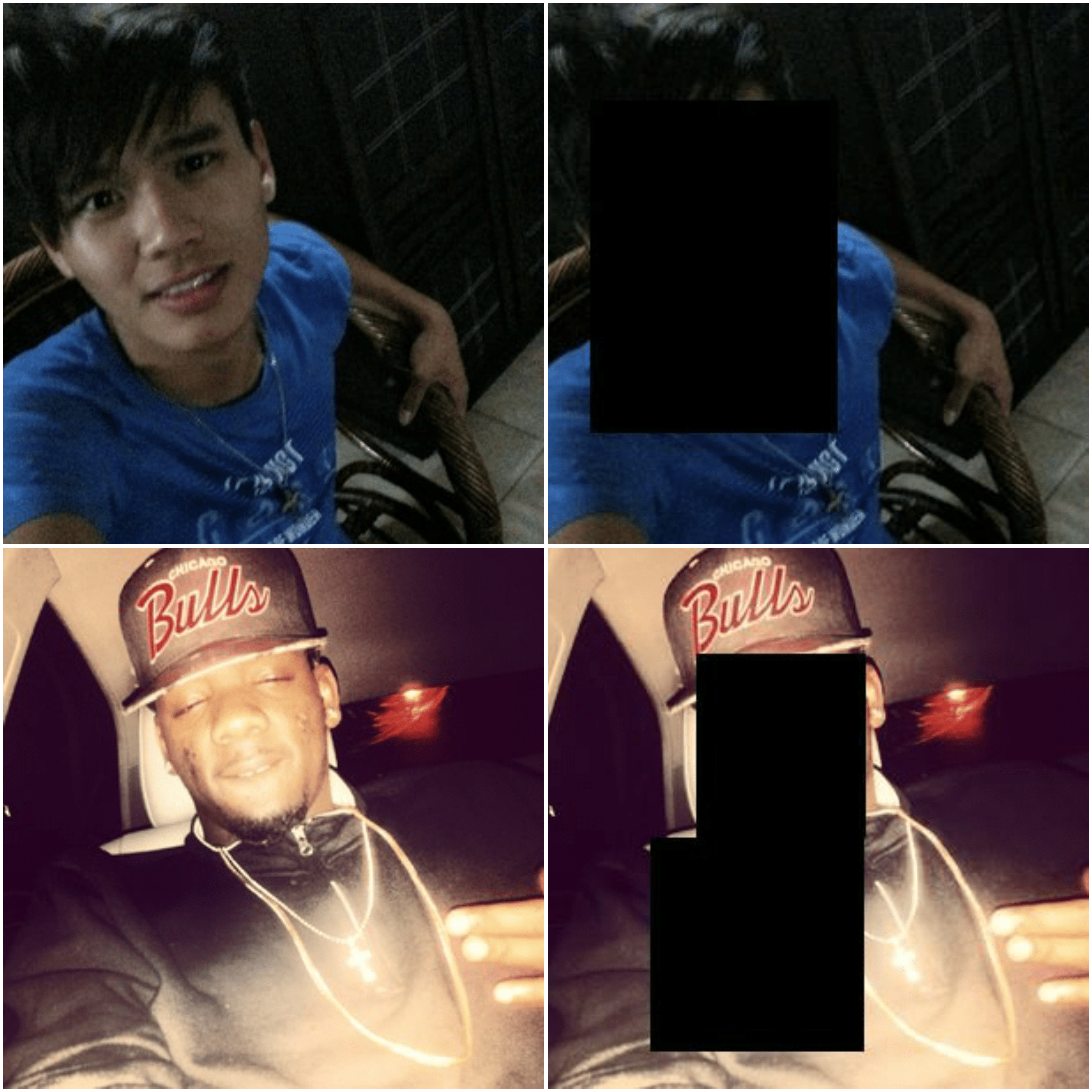}
	}

 		\caption{A few of the original and face-shoulder devoid images.}
 		\label{blur}
  \end{figure}

In order to further analyse the performance of our approach, an ablative comparative study is performed with the other unconstrained CNNs. To ensure that the training process is tractable, and scalable to different problems, a simple experiment is performed to test this fact. With the above trained CNNs, we pass the same testing data to the network, but with a small modification to the input image. We detect the face and shoulder using \cite{zhu2012face}, and set those corresponding pixel values to zero. A subset of $3000$ of these face-shoulder blocked images were chosen and accuracy was checked. A few of these images have been shown in Fig \ref{blur}. Table \ref{blur_accuracy} reports the accuracy in mean Average Precision(mAP) on this experimental setup. It can be seen that the unconstrained network does not show considerable ($<10\%$) drop in accuracy, implying that the networks are learning features which might be potentially from unreliable sources like background, and are neither tractable nor applicable to subtle image analysis problems. However, a considerable drop in accuracy from $86.3\%$ to $58.8\%$ of the proposed constrained approach implies that a separate synergy based training paradigm is enforced and tractable, meaning that the features are learnt from head and shoulder orientation, which is reliable and sensible, and has potential for adaptation to other possible subtle problems. Since the \textit{non-CNN} model of SVM on synergy feature does not yield competitive results in the previous experimental setup, we exclude it in this experimental setting. Besides, tractability is the main focus of this experimental setup, and is more meaningful to test it on the CNN based approaches.
\begin{table}[!t]
	\centering
	\caption{Accuracy on the experimental setup where the input images are devoid of head and shoulders.}
	\label{blur_accuracy}
	\begin{tabular}{|l|l|}
		\hline
		\textbf{Convolutional Networks}                                   & \textbf{\begin{tabular}[c]{@{}l@{}}Accuracy(mAP) on\\ Face and Shoulder\\ Removed Images\end{tabular}} \\ \hline 
	 Unconstrained 	AlexNet \cite{krizhevsky2012imagenet}                                    &    77.1\%                                                                                              \\ \hline
	Unconstrained 	GoogleNet \cite{szegedy2015going}                                  & 75.3\%                                                                                           \\ \hline
		\textbf{Synergy Constrained AlexNet (Proposed)} & \textbf{58.8\%}                                                                                        \\ \hline
	\end{tabular}
\end{table}
\begin{figure*}[!t]
\centering

\begin{subfigure}
 {\includegraphics[width=0.2\textwidth,height=4cm]{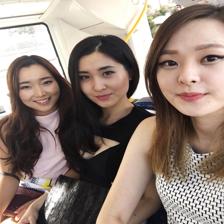}}
 \end{subfigure}
 \begin{subfigure}
  {\includegraphics[width=0.2\textwidth,height=4cm]{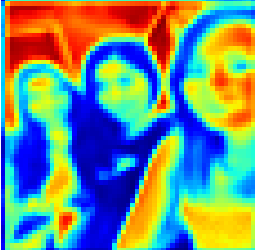}} 
 	\end{subfigure}
 \begin{subfigure}
  {\includegraphics[width=0.2\textwidth,height=4cm]{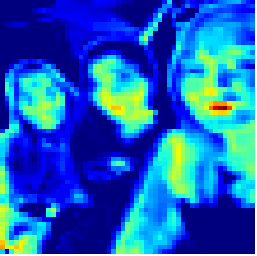}} 
 	\end{subfigure}\\
 		\begin{subfigure}
 		 {\includegraphics[width=0.2\textwidth,height=4cm]{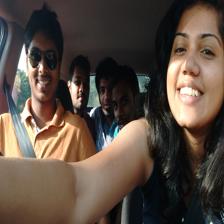}}
 \end{subfigure}
 \begin{subfigure}
  {\includegraphics[width=0.2\textwidth,height=4cm]{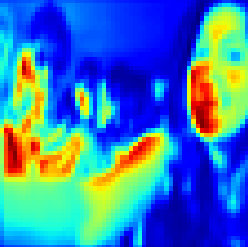}} 
 	\end{subfigure}
 \begin{subfigure}
  {\includegraphics[width=0.2\textwidth,height=4cm]{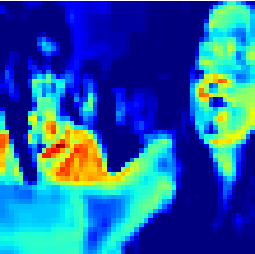}} 
 	\end{subfigure}\\
 	\begin{subfigure}
 	   
 		 {\includegraphics[width=0.2\textwidth,height=4cm]{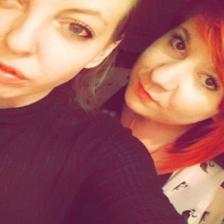}}
 \end{subfigure}
 \begin{subfigure}
 
  {\includegraphics[width=0.2\textwidth,height=4cm]{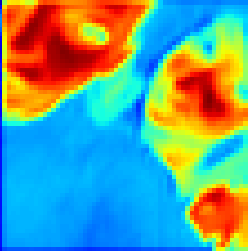}} 
 	\end{subfigure}
 \begin{subfigure}
 
  {\includegraphics[width=0.2\textwidth,height=4cm]{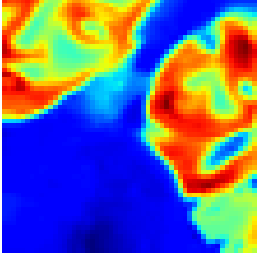}} 
 	\end{subfigure}\\
 	 \begin{subfigure}
 			 
 			 {
 			 \includegraphics[width=0.2\textwidth,height=4cm]{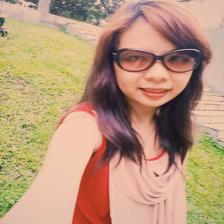}}
 \end{subfigure}
 \begin{subfigure}
 
  {\includegraphics[width=0.2\textwidth,height=4cm]{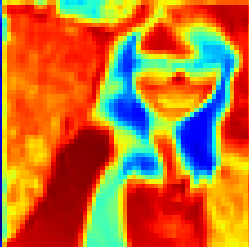}} 
 	\end{subfigure}
 \begin{subfigure}
 
  {\includegraphics[width=0.2\textwidth,height=4cm]{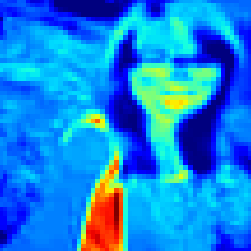}} 
 	\end{subfigure}
 		\caption{ Visualization of different input selfie images (Column I) and its corresponding activations, for selected filters, shown as heat maps for Synergy Constrained AlexNet (Column II) and Unconstrained AlexNet (Column III).}
 		\label{Heatmaps}
  \end{figure*}
\subsubsection{Qualitative Analysis}
In order to determine the features learnt by the networks, the convolutional feature maps are visualized. The Fig. \ref{Heatmaps} illustrates the visualization of different input selfie images and its corresponding activations shown as heat maps for selected filters. First column corresponds to the input RGB selfie image, second column corresponds to the visualization of activations of one of the filters in the first convolutional layer of the proposed constrained network and the third column corresponds to the visualization corresponding to the activations of the same filters  in the first convolutional layer (as in column two) of the the unconstrained but finetuned Alexnet architecture. Although heat maps are shown only for four images, similar activations were observed for other images also, and these four images best summarise the visualizations across the broad range of test data. The heat map of the first selfie from the constrained network concentrates more on the head and shoulder orientation of the person clicking the selfie (selfie-taker) in the image, compared to lower activations of Alexnet. The second image further reinforces the fact that head and shoulder orientation features are being captured in the constrained training paradigm. It is interesting to note here that the activations of the head and shoulder orientation of the primarily selfie-taker is captured, and all the other people in the image are suppressed. It is not the same with the  finetuned Alexnet, where the activations are concentrated upon other people and their shirts. Similar inference can be made from the third image whereas the case having just a single person in the scene has been shown in the fourth image. While our constrained network looks at the head-shoulder alignment more, simple unconstrained Alexnet concentrates on the background which might not be a reliable source of selfie information in most of the cases. Though the visualisations are shown only for one of the filters in each of the networks, this pattern of synergy capture and salient feature learning was consistent in most of the other filters of the first convolutional layers as well.

\section{Conclusion}
\label{conclusion}
This paper presents a synergy constraint based CNN training paradigm for obtaining features which are discriminative for selfie detection. The motivation was drawn based on observations from the way the humans capture the synergy between head orientation and shoulder arm orientation, which was further validated by a small survey. Relevant features that describe these local behaviour were extracted and a synergy measure is obtained by projecting the two features to a common subspace by performing Canonical Correlation Analysis on the two sets of handcraft features. Extensive experimental evaluation by comparing with the existing state of the art architectures was performed to illustrate the importance of having an algorithm based on synergy constraint for selfie detection in particular, and other subtle image analysis problems similar to selfie detection. Although we model a single loss function to constraint the network to learn discriminative features, different losses can be imposed on different layers to better model different synergy relations independently. This might be one of the possible research direction to tackle the problem under consideration. Large scale implementation of synergy based algorithms which involve more than two features for synergy capturing also form part of our future work.



	\bibliographystyle{IEEEtran}
	\bibliography{selfie_cite}

\end{document}